\documentclass[english,communication letter]{IEEEtran}
\usepackage[T1]{fontenc}
\usepackage[latin9]{inputenc}
\usepackage{float}
\usepackage{amsmath}
\usepackage{amssymb}
\usepackage{stackrel}
\usepackage{graphicx}
\usepackage{setspace}

\makeatletter

\providecommand{\tabularnewline}{\\}
\floatstyle{ruled}
\newfloat{algorithm}{tbp}{loa}
\providecommand{\algorithmname}{Algorithm}
\floatname{algorithm}{\protect\algorithmname}

\usepackage[caption=false, font=footnotesize]{subfig}

\usepackage{algorithmic}
\usepackage{cite}

\makeatother

\usepackage{babel}
\begin{document}
\title{Joint User Priority and Power Scheduling for QoS-Aware WMMSE Precoding:
A Constrained-Actor Attentive-Critic Approach\vspace{-0.3cm}
}
\author{\singlespacing{}{\normalsize Kexuan Wang and An Liu, }{\normalsize\textit{Senior Member,
IEEE}}{\normalsize\thanks{Kexuan Wang and An Liu are with the College of Information Science
and Electronic Engineering, Zhejiang University, Hangzhou 310027,
China (email: \{kexuanWang, anliu\}@zju.edu.cn). \textit{(Corresponding
Author: An Liu)}}}\vspace{-0.7cm}
}
\maketitle
\begin{abstract}
6G wireless networks are expected to support diverse quality-of-service
(QoS) demands while maintaining high energy efficiency. Weighted Minimum
Mean Square Error (WMMSE) precoding with fixed user priorities and
transmit power is widely recognized for enhancing overall system performance
but lacks flexibility to adapt to user-specific QoS requirements and
time-varying channel conditions. To address this, we propose a novel
constrained reinforcement learning (CRL) algorithm, Constrained-Actor
Attentive-Critic (CAAC), which uses a policy network to dynamically
allocate user priorities and power for WMMSE precoding. Specifically,
CAAC integrates a Constrained Stochastic Successive Convex Approximation
(CSSCA) method to optimize the policy, enabling more effective handling
of energy efficiency goals and satisfaction of stochastic non-convex
QoS constraints compared to traditional and existing CRL methods.
Moreover, CAAC employs lightweight attention-enhanced Q-networks to
evaluate policy updates without prior environment model knowledge.
The network architecture not only enhances representational capacity
but also boosts learning efficiency. Simulation results show that
CAAC outperforms baselines in both energy efficiency and QoS satisfaction.
\end{abstract}

\begin{IEEEkeywords}
Radio resource allocation, WMMSE, CRL.\vspace{-0.2cm}
\end{IEEEkeywords}

\section{Introduction \label{sec:Introduction}}

With the rise of diverse applications, 6G networks are expected to
efficiently support heterogeneous quality-of-service (QoS) requirements.
For example, delay-sensitive applications such as autonomous driving
and virtual reality require low-latency transmission for safety and/or
immersive experiences, while delay-tolerant applications like web
browsing prioritize high transmission rates \cite{HetNet12}. Meanwhile,
power consumption has also become a critical concern amid the rapid
growth of connected devices and rising environmental pressures \cite{EE0Derrick}.

Weighted Minimum Mean Square Error (WMMSE) is a widely studied technique
enabling MIMO systems to achieve near-optimal throughput, thus improving
overall communication quality \cite{rethinking-WMMSE}. To enhance
its practicality, low-complexity and robust variants were proposed
in \cite{HQY} and \cite{POWMMSE}. However, these methods typically
adopt fixed user priorities and transmit power, neglecting user-specific
QoS requirements and power consumption concerns. Using traditional
iterative optimization methods to address the user priority and transmit
power scheduling (UPTPS) problem would require perfect foresight of
future channel and data traffic, which is unrealistic in highly dynamic
environments. Heuristic approaches in \cite{Qweighted} and \cite{priorityrule1}
mitigate this by prioritizing users with large queues or low past
rates, but lack robustness due to missing theoretical guarantees.

Recently, \cite{model-free-CPO,CMDP2,SoftActorCriticPriority} modeled
radio resource scheduling problems with QoS constraints as Constrained
Markov Decision Processes (CMDPs) and addressed them via Actor-Critic-based
constrained reinforcement learning (CRL). Their Actor module optimizes
the scheduling policy by solving the CMDP, while their Critic module
evaluates policy updates in terms of improving the objective and satisfying
the constraints, ultimately finding a good trade-off between the two
without requiring any environment model knowledge. However, in the
case of UPTPS for WMMSE precoding, the WMMSE precoding process introduces
significant non-convexity into both the objective function and the
QoS constraints, making it difficult for these methods to guarantee
a feasible solution. Specifically, most existing CRL algorithms are
designed for simpler constraints. For example, the widely used trust
region policy optimization-Lagrangian (TRPO-Lag) and proximal policy
optimization-Lagrangian (PPO-Lag) \cite{PPOLagTRPOLag} transform
CMDPs into max-min saddle-point problems and address them by stochastic
gradient descent, which are only suitable for constraints with deterministic
convex feasible sets. The constrained Policy Optimization (CPO) algorithm
\cite{CPO} employs approximate operations to construct convex surrogate
problems with acceptable computational overhead, which also cannot
guarantee a feasible final solution. Moreover, as the number of users
increases, the number of Q-networks grows rapidly, resulting in high
training overhead and slower convergence. Existing methods have largely
overlooked scalability designs for this issue. Our previously proposed
successive convex approximation-based off-policy optimization (SCAOPO)
algorithm \cite{SCAOPO} considers the non-convexity in policy optimization
and employs a simple Monte Carlo (MC) method to avoid Q-networks.
However, its performance degrades with insufficient samples due to
inaccurate evaluation.

To tackle these challenges, we propose a novel Constrained-Actor Attentive-Critic
(CAAC) algorithm to dynamically schedule user priorities and transmit
power for WMMSE. Compared to existing CRL algorithms, CAAC introduces
two key improvements: 1) the Actor module optimizes scheduling policy
by a constrained stochastic successive convex approximation (CSSCA)
method, which can reduces power consumption while effectively satisfying
stochastic non-convex QoS constraints by solving a sequence of convex
surrogate problems; 2) the Critic module employs a lightweight attention-enhanced
Q-network architecture, which significantly improves the training
efficiency and policy evaluation performance. Simulation results show
that CAAC outperforms baseline algorithms in both energy efficiency
and QoS guarantee.

\section{System Model and Problem Formulation}

\subsection{System Model}

\begin{figure}
\begin{centering}
\includegraphics[height=3.1cm]{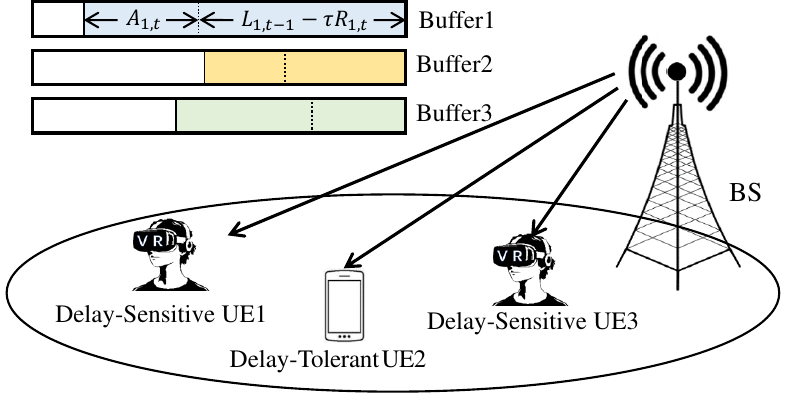}\vspace{-0.3cm}
\par\end{centering}
\caption{\label{fig:System model}Downlink transmission with heterogeneous
QoS support.}
\vspace{-0.5cm}
\end{figure}

This section considers a MIMO system comprising a base station (BS)
equipped with $M$ antennas and $K$ single-antenna users, as depicted
in Fig. \ref{fig:System model}, where the users are indexed by the
set $\mathcal{K}=\bigl\{1,\ldots,K\bigr\}$. The time dimension is
divided into timeslots of duration $\tau$, each indexed by $t$.
Moreover, the channel from BS to the $k$-th user is denoted by $\boldsymbol{h}_{k,t}\in\mathbb{C}^{1\times M},\forall t.$

We now introduce the downlink transmission process. At each timeslot
$t$, bursty data of length $A_{k,t}$ is generated with probability
$P_{k}$ and stored in a dedicated buffer for user $k$, along with
any previously untransmitted data. Specifically, the queue length
is updated according to\vspace{-0.2cm}
\[
L_{k,t}=L_{k,t-1}-R_{k,t-1}\tau+A_{k,t},\forall k,t,
\]
where $R_{k,t-1}$ is the downlink transmission rate for user $k$
at timeslot $t-1$. Based on the current channel conditions and data
queue states, the BS assigns user priority weights $\boldsymbol{\omega}_{t}\triangleq\left\{ \omega_{k,t}\right\} _{k\in\mathcal{K}}$
and the total transmit power $p_{t}$, and then executes the WMMSE
precoding \cite{rethinking-WMMSE}. Then, the obtrained precoder vector
$\boldsymbol{v}_{k,t}$ is used to process the transmit signal $s_{k,t}$
for user $k$, where $\sum_{k=1}^{K}\mathrm{Tr}\left(\boldsymbol{v}_{k,t}^{H}\boldsymbol{v}_{k,t}\right)=p_{t}$
and $\left|s_{k,t}\right|=1$. Further, the received signal $y_{k,t}\in\mathbb{C}$
at user $k$ is given by\vspace{-0.2cm}
\[
y_{k,t}=\boldsymbol{h}_{k,t}\boldsymbol{v}_{k,t}s_{k,t}+\sum_{m\neq k}^{K}\boldsymbol{h}_{k,t}\boldsymbol{v}_{m,t}s_{m,t}+n_{k},\forall k,t
\]
where $n_{k}\sim\mathcal{CN}\bigl(0,\sigma_{k}^{2}\bigr)$ represents
the additive white Gaussian noise. Finally, the downlink transmission
rate for user $k$ at the current timeslot $t$ is expressed as\vspace{-0.2cm}
\[
R_{k,t}=W\mathrm{log}_{2}\Bigl(1+\frac{\bigl|\bigl(\boldsymbol{h}_{k,t}\bigr)^{H}\boldsymbol{v}_{k,t}\bigr|^{2}}{\sum_{m\neq k}\bigl|\bigl(\boldsymbol{h}_{k,t}\bigr)^{H}\boldsymbol{v}_{m,t}\bigr|^{2}+\sigma_{k}^{2}}\Bigr),
\]
where $W$ denotes the bandwidth.

\subsection{Problem Formulation}

This paper aims to find the optimal UPTPS for WMMSE precoding, which
minimizes the system energy consumption under heterogeneous QoS constraints.
This can be formulated as the following sequential decision-making
problem:\vspace{-0.2cm}
\begin{align}
\underset{\left\{ \boldsymbol{\omega}_{t},p_{t}\right\} _{t=0,1,\ldots}}{\mathrm{max}} & \underset{T\rightarrow\infty}{\mathrm{lim}}\frac{1}{T}\mathbb{E}\biggl[\stackrel[t=0]{T-1}{\sum}p_{t}\biggr]\label{eq:original-problem}\\
\mathrm{s.t.}\underset{T\rightarrow\infty}{\mathrm{lim}} & \frac{1}{T}\mathbb{E}\biggl[\stackrel[t=0]{T-1}{\sum}\stackrel[n=1]{N}{\sum}\delta\left(n-\varrho_{k}\right)\mathbb{U}\left(n,k,t\right)\biggr]\leq c_{k},\forall k.\nonumber 
\end{align}
where $\mathbb{E}\left[\cdot\right]$ is the expectation taken over
channel fading distributions and data arrival processes, $\varrho_{k}$
represents the QoS category required by user $k$, $\delta\left(\cdot\right)$
is discrete impulse response function, $\mathbb{U}\left(n,k,t\right)$
denotes the utility function corresponding to the $n$-th QoS type,
and $c_{b,k}$ denotes the minimum QoS threshold acceptable for user
$k$.

To accommodate diverse applications, the system is required to support
$N\geqslant1$ types of QoS demands. This paper considers two type
of QoS requirements as representative examples: 1) If the $k$-th
user is delay-sensitive, we define the utility function $\mathbb{U}\left(1,k,t\right)=L_{k,t}/a_{k}$,
where $a_{k}=P_{k,t}\mathbb{E}\left[A_{k,t}\right]$ denotes the average
data arrival per timeslot for user $k$; 2) If the $k$-th user is
delay-tolerant, the utility function is defined as $\mathbb{U}\left(2,k,t\right)=-\mathcal{R}_{k,t}$.
It is worth noting that additional QoS requirements of practical interest
can also be incorporated by redefining the corresponding utility functions
$\mathbb{U}\left(n,k,t\right)$, without requiring any modification
to the proposed algorithm. 

\section{The Constrained-Actor Attentive-Critic Algorithm Design}

\subsection{Constrained Markov Decision Process}

To facilitate tractable analysis, we model the original problem (\ref{eq:original-problem})
as a Constrained Markov Decision Process (CMDP), represented by a
tuple $\left(\mathcal{S},\mathcal{\mathcal{A}},P,R,C\right)$, where:
\begin{itemize}
\item \textbf{The state space} $\mathcal{S}$\textbf{: }the state at timeslot
$t$ is $\boldsymbol{s}_{t}\triangleq\bigl\{\mathbf{L}_{t};\mathbf{H}_{t}\bigr\}$,
where $\mathbf{L}_{t}\triangleq\left[L_{1,t},L_{2,t},\ldots,L_{K,t}\right]^{\intercal}$
denotes the queue length vector, and $\mathbf{H}_{t}\triangleq\left[\boldsymbol{h}_{1,t},\boldsymbol{h}_{2,t},\ldots,\boldsymbol{h}_{K,t}\right]^{\intercal}$
is the channel state information (CSI). Accordingly, the state space\textbf{
}$\mathcal{S}$ is defined as the Cartesian product of the queue length
space $\mathbb{R}_{+}^{K}$ and the CSI space $\mathbb{C}^{K\times M}$.
\item \textbf{The action space} $\mathcal{A}$\textbf{:} the action $\boldsymbol{a}_{t}\triangleq\bigl\{\boldsymbol{\omega}_{t},p_{t}\bigr\}$
is sampled from the action space $\mathcal{A}$ according to a policy
$\pi:\mathcal{S}\rightarrow\mathbf{P}\left(\mathcal{A}\right)$, where
$\pi\left(\boldsymbol{a}_{t}\mid\boldsymbol{s}_{t}\right)$ denoting
the probability of selecting $\boldsymbol{a}_{t}$ given state $\boldsymbol{s}_{t}$.
To mitigate the curse of dimensionality, the policy $\pi$ is parameterized
by deep neural networks (DNNs), and is further denoted by $\pi_{\boldsymbol{\theta}}$.
\item \textbf{The transition probability function }$P$\textbf{:} the environment
evolves according to an unknown transition probability function $P:\mathcal{S}\times\mathcal{A}\times\mathcal{S}\rightarrow\left[0,1\right]$,
where $P\left(\boldsymbol{s}_{t+1}\mid\boldsymbol{s}_{t},\boldsymbol{a}_{t}\right)$
denotes the probability of reaching state $\boldsymbol{s}_{t+1}$
from $\boldsymbol{s}_{t}$ given action $\boldsymbol{a}_{t}$. Together,
the function $P$ and policy $\pi_{\boldsymbol{\theta}}$ determine
a trajectory distribution $\sigma_{\pi}$, i.e., $\boldsymbol{s}_{t}\thicksim P\left(\cdot\mid\boldsymbol{s}_{t-1},\boldsymbol{a}_{t-1}\right),\boldsymbol{a}_{t}\thicksim\pi_{\boldsymbol{\theta}}\left(\cdot\mid\boldsymbol{s}_{t}\right)$.
\item \textbf{Reward function $R$ and $C$:} after taking action $\boldsymbol{a}_{t}$,
the agent (BS) obtains a system throughput-related reward $R\left(\boldsymbol{s}_{t},\boldsymbol{a}_{t}\right)\triangleq p_{t}$,
and some QoS-related rewards $C_{k}\left(\boldsymbol{s}_{t},\boldsymbol{a}_{t}\right)\triangleq\stackrel[n=1]{N}{\sum}\delta\left(n-\varrho_{k}\right)\mathbb{U}\left(n,k,t\right)$.
\end{itemize}
Accordingly, the original problem (\ref{eq:original-problem}) can
be reformulated as:\vspace{-0.2cm}
\begin{align}
\underset{\theta\in\Theta}{\mathrm{min}}f_{0}\left(\boldsymbol{\theta}\right)\overset{\triangle}{=} & \underset{T\rightarrow\infty}{\mathrm{lim}}\frac{1}{T}\mathbb{E}_{p_{\pi_{\boldsymbol{\theta}}}}\biggl[\stackrel[t=0]{T-1}{\sum}C_{0}^{'}\left(\boldsymbol{s}_{t},\boldsymbol{a}_{t}\right)\biggr]\label{eq:CMDP-problem}\\
\mathrm{s.t.}f_{k}\left(\boldsymbol{\theta}\right)\overset{\triangle}{=} & \underset{T\rightarrow\infty}{\mathrm{lim}}\frac{1}{T}\mathbb{E}_{p_{\pi_{\boldsymbol{\theta}}}}\biggl[\stackrel[t=0]{T-1}{\sum}C_{k}^{'}\left(\boldsymbol{s}_{t},\boldsymbol{a}_{t}\right)\biggr]\leq0,\forall k,\nonumber 
\end{align}
where $C_{0}^{'}\left(\boldsymbol{s}_{t},\boldsymbol{a}_{t}\right)=R\left(\boldsymbol{s}_{t},\boldsymbol{a}_{t}\right)$
and $C_{k}^{'}\left(\boldsymbol{s}_{t},\boldsymbol{a}_{t}\right)=C_{k}\left(\boldsymbol{s}_{t},\boldsymbol{a}_{t}\right)-c_{k},k=1,2,\ldots,K$.
In the following, we propose the Constrained-Actor Attentive-Critic
(CAAC) algorithm, which alternates between Actor and Critic modules
to solve this CMDP. The algorithm design is detailed below, with the
overall procedure summarized in Algorithm 1.
\begin{algorithm}[t]
\caption{\label{alg:CAAC} Constrained-Actor Attentive-Critic Algorithm}

\textbf{Input:} The initial entries of $\kappa_{1},\kappa_{2},\kappa_{3}$,
$\boldsymbol{\omega}_{0}$ and $\boldsymbol{\theta}_{0}$.

\textbf{for} $t=0,1,2\cdots$

\phantom{} \phantom{}\textbf{ }Sample the new observation $\tilde{\varepsilon}_{t}$
and update the set $\varepsilon_{i}$.

\phantom{} \phantom{}\textbf{ }\textbf{\textit{The Critic Module}}\textbf{:}

\phantom{} \phantom{}\textbf{ }\phantom{} \phantom{} Calculate
$\boldsymbol{G}_{i,t_{\mathrm{cri}}}^{\mathcal{B}}$ and update $\boldsymbol{\omega}^{i}$
by (\ref{eq:TD-updates}) for $T_{\mathrm{cri}}$ times.

\phantom{} \phantom{}\textbf{ }\textbf{\textit{The Actor Module}}\textbf{:}

\phantom{} \phantom{}\textbf{ }\phantom{} \phantom{} Estimate
$\hat{f}_{k,i}$ and $\hat{\boldsymbol{g}}_{k,i}$ according to (\ref{eq:f-tilde})-(\ref{eq:g-hat}).

\phantom{} \phantom{}\textbf{ }\phantom{} \phantom{} Update the
surrogate function $\left\{ \bar{f}_{k,t}\left(\boldsymbol{\theta}\right)\right\} _{k=0,\ldots,K}$via
(\ref{eq:surrogate functions}).

\phantom{} \phantom{}\textbf{ }\phantom{} \phantom{} \textbf{if
}Problem (\ref{eq:objective update}) is feasible: Solve (\ref{eq:objective update})
to obtain $\bar{\boldsymbol{\theta}}_{t}$.

\phantom{} \phantom{}\textbf{ }\phantom{} \phantom{} \textbf{else}:
Solve (\ref{eq:feasible update}) to obtain $\bar{\boldsymbol{\theta}}_{t}$.
\textit{(Feasible update)}

\phantom{} \phantom{}\textbf{ }\phantom{} \phantom{} Update policy
parameters $\boldsymbol{\theta}_{t+1}$ according to (\ref{eq:theta update}).
\end{algorithm}
\vspace{-0.3cm}

\subsection{The Actor Module}

The Actor module executes the policy network and optimizes it using
CSSCA method. CSSCA estimates the values and gradients of the objective
and constraint functions through interactions with the environment,
and then handle the non-convex CMDP by solving a sequence of convex
quadratic optimization problems constructed from these estimates.

At the beginning of the $i$-th policy update, the policy $\pi_{\boldsymbol{\theta}_{i}}$
interacts with the environment $B$ times, collecting an observation
set $\varepsilon_{i}=\bigl\{\tilde{\varepsilon}_{t}\bigr\}_{t=Bi-B+1:Bi}$,
where each tuple is given by $\tilde{\varepsilon}_{t}\triangleq\bigl\{\boldsymbol{s}_{t},\boldsymbol{a}_{t},\bigl\{ C_{l}^{\text{'}}(\boldsymbol{s}_{t},\boldsymbol{a}_{t})\bigr\}_{l=0,\ldots,L},\boldsymbol{s}_{t+1}\bigr\}$.
These tuples are stored in a replay buffer for function and gradient
estimation. Specifically, first apply the sample average approximation
(SAA) method to compute sample-based estimates of the objective and
constraint functions:
\begin{equation}
\tilde{f}_{k,i}=\hat{\mathbb{E}}_{\varepsilon_{i}}\bigl[C_{k,t}^{\text{'}}\bigr],\forall k,t.\label{eq:f-tilde}
\end{equation}
where $\hat{\mathbb{E}}_{\varepsilon_{i}}$ denotes the empirical
mean over the observation set $\varepsilon_{i}$. Moreover, according
to the policy gradient theorem \cite{SCAOPO}, the gradient $\bigtriangledown f_{k}\left(\boldsymbol{\theta}\right)$
is expressed as 
\[
\nabla f_{k,i}\left(\boldsymbol{\theta}\right)=\mathbb{E}\left[Q_{k}^{\pi_{\boldsymbol{\theta}_{i}}}\left(\boldsymbol{s},\boldsymbol{a}\right)\nabla_{\boldsymbol{\theta}}\textrm{log}\pi_{\boldsymbol{\theta}_{i}}\left(\boldsymbol{a}\mid\boldsymbol{s}\right)\right],\forall k,
\]
where $\bigl\{ Q_{k}^{\pi_{\boldsymbol{\theta}_{i}}}\bigr\}_{k=0,\ldots,K}$
serve as crucial tools for assessing the effectiveness of the current
policy in improving system throughput or satisfying QoS constraints,
and are defined as
\begin{align*}
Q_{k}^{\pi_{\boldsymbol{\theta}_{i}}}\left(\boldsymbol{s},\boldsymbol{a}\right)= & \mathbb{E}\Bigl[\sum_{l=0}^{\infty}\Bigl(C_{k,l}^{\text{'}}-f_{k}\left(\boldsymbol{\theta}_{i}\right)\Bigr)\bigl|\boldsymbol{s}_{0},\boldsymbol{a}_{0}=\boldsymbol{s},\boldsymbol{a}\Bigr].
\end{align*}
These Q-functions are approximated by Q-networks $\bigl\{ Q_{k}^{\boldsymbol{\omega}_{i}}\bigr\}_{k=0,\ldots,K}$
in the Critic module. Similarly, following the SAA, the sample-based
policy gradients are estimated as
\begin{align}
\tilde{\boldsymbol{g}}_{k,i}= & \hat{\mathbb{E}}_{\varepsilon_{i}}\left[Q_{k}^{\boldsymbol{\omega}_{i}}\left(\boldsymbol{s},\boldsymbol{a}\right)\nabla_{\boldsymbol{\theta}}\textrm{log}\pi_{\boldsymbol{\theta}_{i}}\left(\boldsymbol{a}\mid\boldsymbol{s}\right)\right],\forall k.\label{eq:g-tilde}
\end{align}
To stabilize the learning process, we apply recursive updates to the
function and gradient estimates as follows:
\begin{equation}
\hat{f}_{k,i+1}=\left(1-\eta_{i+1}\right)\hat{f}_{k,i}+\eta_{i}\tilde{f}_{k,i},\forall k,i,\label{eq:f-hat}
\end{equation}
\begin{equation}
\hat{\boldsymbol{g}}_{k,i+1}=\left(1-\eta_{i+1}\right)\hat{\boldsymbol{g}}_{k,i}+\eta_{i}\tilde{\boldsymbol{g}}_{k,i},\forall k,i,\label{eq:g-hat}
\end{equation}
where $\eta_{i}\triangleq\bigl(i+1\bigr)^{-\kappa_{1}}$ and $\kappa_{1}\in\bigl(0,1\bigr)$.

Using $\hat{f}_{k,i}$ and $\hat{\boldsymbol{g}}_{k,i}$, we construct
convex quadratic surrogate functions for the objective/constraint
functions as follows:\vspace{-0.2cm}
\begin{equation}
\bar{f}_{k,i}\left(\boldsymbol{\theta}\right)=\hat{f}_{k,i}+\left(\hat{\boldsymbol{g}}_{k,i}\right)^{\textrm{\ensuremath{\intercal}}}\left(\boldsymbol{\theta}-\boldsymbol{\theta}_{i}\right)+\zeta_{k}\left\Vert \boldsymbol{\theta}-\boldsymbol{\theta}_{i}\right\Vert _{2}^{2},\forall k\label{eq:surrogate functions}
\end{equation}
where $\zeta_{k}>0$ is a constant. The policy update is then carried
out by solving the following objective optimization problem:\vspace{-0.2cm}
\begin{align}
\bar{\boldsymbol{\theta}}_{i}=\underset{\boldsymbol{\theta}\in\mathbf{\Theta}}{\textrm{argmin}}\  & \bar{f}_{0,i}\left(\boldsymbol{\theta}\right)\ \ \ \ \textrm{s.t.}\ \bar{f}_{k,i}\left(\boldsymbol{\theta}\right)\leq0,\forall k.\label{eq:objective update}
\end{align}
If it is infeasible, we solve a feasible optimization problem:\vspace{-0.2cm}
\begin{align}
\bar{\boldsymbol{\theta}}_{i}=\underset{\boldsymbol{\theta}\in\mathbf{\Theta},y}{\textrm{argmin}}\  & \alpha\ \ \ \ \textrm{s.t.}\ \bar{f}_{k,i}\left(\boldsymbol{\theta}\right)\leq\alpha,\forall k.\label{eq:feasible update}
\end{align}
Both (\ref{eq:objective update}) and (\ref{eq:feasible update})
can be easily solved by standard convex optimization algorithms, e.g.
Lagrange-dual methods. Finally, the policy parameters are updated
via:
\begin{equation}
\boldsymbol{\theta}_{i}=(1-\mu_{i})\boldsymbol{\theta}_{i-1}+\mu_{i}\bar{\boldsymbol{\theta}}_{i}.\label{eq:theta update}
\end{equation}
where $\mu_{i}\triangleq\bigl(i+1\bigr)^{-\kappa_{2}}$ and $\kappa_{2}\in\bigl(0,1\bigr)$.
\vspace{-0.3cm}

\subsection{The Critic Module}

The Critic module aims to train Q-networks $\bigl\{ Q_{k}^{\boldsymbol{\omega}_{i}}\bigr\}_{k=0,\ldots,K}$
to approximate Q-functions $\bigl\{ Q_{k}^{\pi_{\boldsymbol{\theta}_{i}}}\bigr\}_{k=0,\ldots,K}$.

\subsubsection{Lightweight Attention-Enhanced Q-Networks}

\begin{figure}
\begin{centering}
\includegraphics[height=2.9cm]{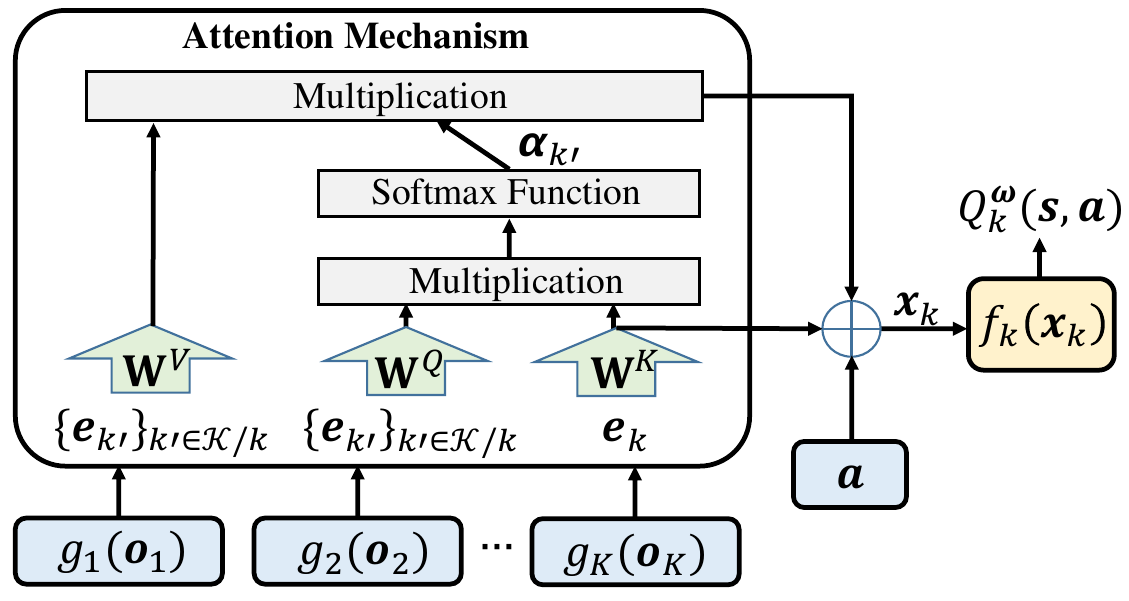}\vspace{-0.3cm}
\par\end{centering}
\caption{\label{fig:Q-networks}The lightweight attention-enhanced Q-network
architecture.}
\vspace{-0.5cm}
\end{figure}

As shown in Fig. \ref{fig:Q-networks}, we designs a lightweight attention-enhanced
architecture for Q-networks, comprising three main components:
\begin{itemize}
\item \textbf{A Shared Embedding Layer: }The embedding layer is composed
of a set of embedding functions $\left\{ g_{k}\bigl(\cdot\bigr)\right\} _{k=1:K}$,
which are shared across all Q-networks. We decompose the state $\boldsymbol{s}_{t}$
and priority weights $\boldsymbol{\omega}_{t}$ into some user-specific
tuples $\left\{ \boldsymbol{o}_{k,t}\triangleq\bigl(L_{k,t},\boldsymbol{h}_{k,t},\omega_{k,t}\bigr)\right\} _{k=1}^{K}$.
Each tuple $\boldsymbol{o}_{k,t}$ can be encoded into a low-dimensional
embedding vector $\boldsymbol{e}_{k,t}\triangleq g_{k}\bigl(\boldsymbol{o}_{k,t}\bigr)$
through $g_{k}\bigl(\cdot\bigr)$.
\item \textbf{A Shared Attention Layer: }Considering that the QoS performance
of a user $k$ is primarily determined by the action $\boldsymbol{a}_{t}$
and its own state $\boldsymbol{o}_{k,t}$, but may also be affected
by neighboring users to varying degrees, we construct a feature vector
$\boldsymbol{x}_{k,t}$ for each Q-network $k$ as follows:\vspace{-0.2cm}
\begin{equation}
\boldsymbol{x}_{k,t}=\boldsymbol{a}_{t}\oplus\boldsymbol{v}_{k,t}\oplus\sum_{k'\in\mathcal{K}/k}\alpha_{k,k',t}\boldsymbol{v}_{k',t},
\end{equation}
where the value vector$\boldsymbol{v}_{k',t}=\mathbf{W}^{V}\boldsymbol{e}_{k',t}$
is a linear projection of $\boldsymbol{e}_{k',t}$, $\alpha_{k,k',t}$
denotes the attention weight, and $\oplus$ denotes vector concatenation.
For any $k\in\mathcal{K}$, the attention weight $\alpha_{k,k',t}$
is computed according to\vspace{-0.2cm}
\begin{equation}
\alpha_{k,k',t}=\mathrm{softmax}\bigl(\bigl(\boldsymbol{k}_{k,t}\bigr)^{T}\boldsymbol{q}_{k',t}\bigr),
\end{equation}
where $\boldsymbol{k}_{k,t}\triangleq\mathbf{W}^{K}\boldsymbol{e}_{k,t}$
and $\boldsymbol{q}_{k',t}\triangleq\mathbf{W}^{Q}\boldsymbol{e}_{k',t}$
are the key and query vectors obtained via linear transformations.
The inner product $\bigl(\boldsymbol{k}_{k,t}\bigr)^{T}\boldsymbol{q}_{k',t}$
reflects the relevance between user $k$' state and user $k'$' state.
The parameter matrices $\mathbf{W}^{K}\in\mathbb{R}^{L_{1}\times L_{2}/2}$,
$\mathbf{W}^{Q}\in\mathbb{R}^{L_{1}\times L_{2}/2}$, and $\mathbf{W}^{V}\in\mathbb{R}^{L_{1}\times L_{2}/2}$
are shared across all Q-networks. In the case of $k=0$, $\boldsymbol{v}_{k,t}$
degenerates into an empty vector, and the attention weights are set
uniformly (i.e., $\alpha_{0,k',t}=1$), i.e., the Q-network corresponding
to objective function equally attends to each user' state.
\item \textbf{Separate Output Functions:} Finally, each $\boldsymbol{x}_{k,t}$
is passed through a separate output function $f_{k}\bigl(\cdot\bigr)$,
a two-layer fully connected network (FCN), to produce $Q_{k}^{\boldsymbol{\omega}_{i}}\bigl(\boldsymbol{s}_{t},\boldsymbol{a}_{t}\bigr)$.
\end{itemize}
Conventional Actor-Critic-based CRL typically uses independent Q-networks
for each Q-function. In contrast, Q-networks in this work adopts shared
embedding and attention layers, reducing the parameter size and improving
training efficiency. Since all users share the same structure of state
information, this shared design encourages a common embedding space,
thereby enhancing generalization. Moreover, the attention mechanism
enables Q-networks to focus on critical information, boosting their
representational capacity. 

\subsubsection{The training for Q-networks}

Then, the Q-networks $\bigl\{ Q_{k}^{\boldsymbol{\omega}_{i}}\bigr\}_{k=0,\ldots,K}$
are optimized according to the classical temporal-difference (TD)-learning
method \cite{DQlearning}, which minimizes the Bellman error of the
Q-networks:\vspace{-0.2cm}
\begin{align*}
\mathcal{B}_{k}^{\boldsymbol{\omega}_{i}}\bigl(\boldsymbol{s}_{t},\boldsymbol{a}_{t},\boldsymbol{s}_{t+1}\bigr)= & \sum_{k=0}^{K}\bigl|Q_{k}^{\boldsymbol{\omega}_{i}}\left(\boldsymbol{s}_{t},\boldsymbol{a}_{t}\right)-\mathcal{T}Q_{k}^{\boldsymbol{\omega}_{i}}\left(\boldsymbol{s}_{t},\boldsymbol{a}_{t}\right)\bigr|^{2},
\end{align*}
where $\mathcal{T}$ is the Bellman operator defined as\vspace{-0.2cm}
\[
\mathcal{T}Q_{k}^{\boldsymbol{\omega}_{i}}\left(\boldsymbol{s}_{t},\boldsymbol{a}_{t}\right)=\mathbb{E}\Bigl[Q_{\boldsymbol{\omega}_{k,i}}\left(\boldsymbol{s}_{t+1},\boldsymbol{a}_{t+1}'\right)+C_{k,t}^{\text{'}}\Bigr]-f_{k}\left(\boldsymbol{\theta}_{i}\right),
\]
where $\boldsymbol{a}_{t+1}'$ is the action chosen by $\pi_{\boldsymbol{\theta}_{i}}$
at the next state $\boldsymbol{s}_{t+1}$. Since it is unrealistic
to obtain the accurate function value $f_{k}\left(\boldsymbol{\theta}_{i}\right)$
online, we replace it by $\hat{f}_{k,i}$. We divide the new observation
set $\varepsilon_{i}$ into $T_{\mathrm{cri}}$ mini-batches (with
$T_{\mathrm{cri}}<B$, where $B$ is the total number of new observation
samples in $\varepsilon_{i}$), indexed by $\bigl[\varepsilon_{i}^{1},\ldots,\varepsilon_{i}^{t_{\mathrm{cri}}},\ldots,\varepsilon_{i}^{T_{\mathrm{cri}}}\bigr]$,
and perform the following TD-updates for $T_{\mathrm{cri}}$ times:\vspace{-0.2cm}
\begin{equation}
\boldsymbol{\omega}_{i}\leftarrow\boldsymbol{\omega}_{i}-\upsilon_{i}\boldsymbol{G}_{i,t_{\mathrm{cri}}}^{\mathcal{B}},\forall k,t_{\mathrm{cri}},\label{eq:TD-updates}
\end{equation}
where $\boldsymbol{G}_{i,t_{\mathrm{cri}}}^{\mathcal{B}}$ is a stochastic
gradient term of Bellman error:\vspace{-0.2cm}
\begin{align}
\boldsymbol{G}_{i,t_{\mathrm{cri}}}^{\mathcal{B}}= & \sum_{t\in\varepsilon_{i}^{t_{\mathrm{cri}}}}\Bigl(Q_{k}^{\boldsymbol{\omega}_{i}}\left(\boldsymbol{s}_{t},\boldsymbol{a}_{t}\right)-C_{k,t}^{\text{'}}+\hat{f}_{k,i}\nonumber \\
-Q_{\boldsymbol{\omega}_{k,i}} & \bigl(\boldsymbol{s}_{t+1},\boldsymbol{a}_{t+1}'\bigr)\Bigr)\nabla_{\boldsymbol{\omega}}Q_{k}^{\boldsymbol{\omega}_{i}}\left(\boldsymbol{s}_{t},\boldsymbol{a}_{t}\right),\forall k,\label{eq:gradient w}
\end{align}
where the step size $\upsilon_{i}\triangleq\bigl(i+1\bigr)^{-\kappa_{3}}$
with $\kappa_{3}\in\bigl(0,1\bigr)$. \vspace{-0.3cm}

\subsection{Convergence Analysis}

According to our previous work \cite[Lemma 3]{SCAOPO}, the design
of the step size $\eta_{i}$ ensures that $\hat{f}_{k,i}$ asymptotically
converges to $f_{k}\left(\boldsymbol{\theta}_{i}\right)$. Then, the
TD updates can ensure that the Q-networks accurately approximate the
Q-functions, as long as the representations of the Q-networks are
strong enough and both the sample size $B$ and the number of TD updates
$T_{\mathrm{cri}}$ are sufficiently large \cite{DQlearning}. In
this case, $\tilde{\boldsymbol{g}}_{k}^{t}$ is an unbias estimation
of $\nabla J_{k}\left(\boldsymbol{\theta}_{t}\right)$, and $\hat{\boldsymbol{g}}_{k,i}$
can also asymptotically converge to $\nabla J_{k}\left(\boldsymbol{\theta}_{t}\right)$,
provided that the step size parameters satisfy $\kappa_{1}<\kappa_{2}$
\cite[Lemma 3]{SCAOPO}. As iterations proceed, the objective updates
(\ref{eq:objective update}) make the value of the objective function
increase continuously and finally achieve a stationary point of the
original problem (\ref{eq:CMDP-problem}); the feasible updates (\ref{eq:feasible update})
ensure the iteration point within or not far from the feasible QoS
region, and asymptotically converges into the feasible QoS region
almost surely. In practice, our lightweight network design enables
the Q-networks to quickly converge to a small approximation error,
thereby exhibiting good convergence performance despite the limited
$B$ and $T_{\mathrm{cri}}$. This conclusion is also supported by
simulation results in Section \ref{sec:Numerical-Results}.

\section{Numerical Results \label{sec:Numerical-Results}}

We consider a 500-meter radius cell with a BS equipped with $M=16$
antennas serving $K\in\bigl\{4,8,12,16\bigr\}$ single-antenna users.
Each user moves in a random direction at 3 km/h and reverses direction
at the cell boundary. The system operates at a carrier frequency of
$f_{0}=3.5$ GHz with a timeslot duration of $\tau=1$ ms, transmission
bandwidth $W=1$ MHz, and noise power density $\delta_{0}=-174$ dBm/Hz.
A block fading channel model $\boldsymbol{h}_{k,t}=\sqrt{\alpha_{k,t}}\tilde{\boldsymbol{h}}_{k,t}$
is used following \cite{MARL0RRS1}, where $\alpha_{k,t}=34+40\mathrm{lg}\bigl(d_{k}\bigr)$
models large-scale fading, and $d_{k}$ is the distance between the
BS and user $k$. Small-scale fading follows the Jakes model $\tilde{\boldsymbol{h}}_{k,t}=\rho\tilde{\boldsymbol{h}}_{k,t-t}+\sqrt{1-\rho^{2}}\boldsymbol{u}_{k,t}$,
where $\tilde{\boldsymbol{h}}_{k,0}$and $\boldsymbol{u}_{k,t}$ are
i.i.d. complex Gaussian variables with unit variance, and $\rho=J_{0}\left(2\pi f_{d}\tau\right)$
is the correlation coefficient with $J_{0}$ being the zeroth-order
Bessel function and $f_{d}$ the maximum Doppler frequency. Half of
the users are delay-sensitive, with $P_{k}\in\bigl[0.4,0.6\bigr]$
and $A_{k,t}$ sampled from distribution $P\bigl(A_{k,t}=x\bigr)=\frac{\left(A_{k,t}\right)^{x}}{x!}e^{-\lambda_{k}}$,
where $\lambda_{k}\in\bigl[5,15\bigr]$ Kbits. The maximum acceptable
average delay for them is assumed to be $c_{k,b}=3$ timeslots. The
remaining users are delay-tolerant, with a minimum transmission rate
$c_{k,b}=5$ Mbps. The policy employs a three-layer FCN with a hidden
layer width of 256. The embedding and attention layers of Q-networks
have output dimensions $L_{1}K$ ($L_{1}=2$) and $L_{2}=64$, while
the output functions' hidden layer has $L_{3}=32$ units. At each
iteration, the agent collects $B=200$ observations and updates Q-networks
for $T_{\mathrm{cri}}=10$ times. Step sizes are $\alpha_{t}=\frac{1}{t^{0.6}}$,
$\beta_{t}=\frac{1}{t^{0.7}}$, and $\gamma_{t}=\frac{1}{t^{0.3}}$.
All results are averaged over 5 random seeds. 
\begin{figure}
\centering{}\includegraphics[height=4.2cm]{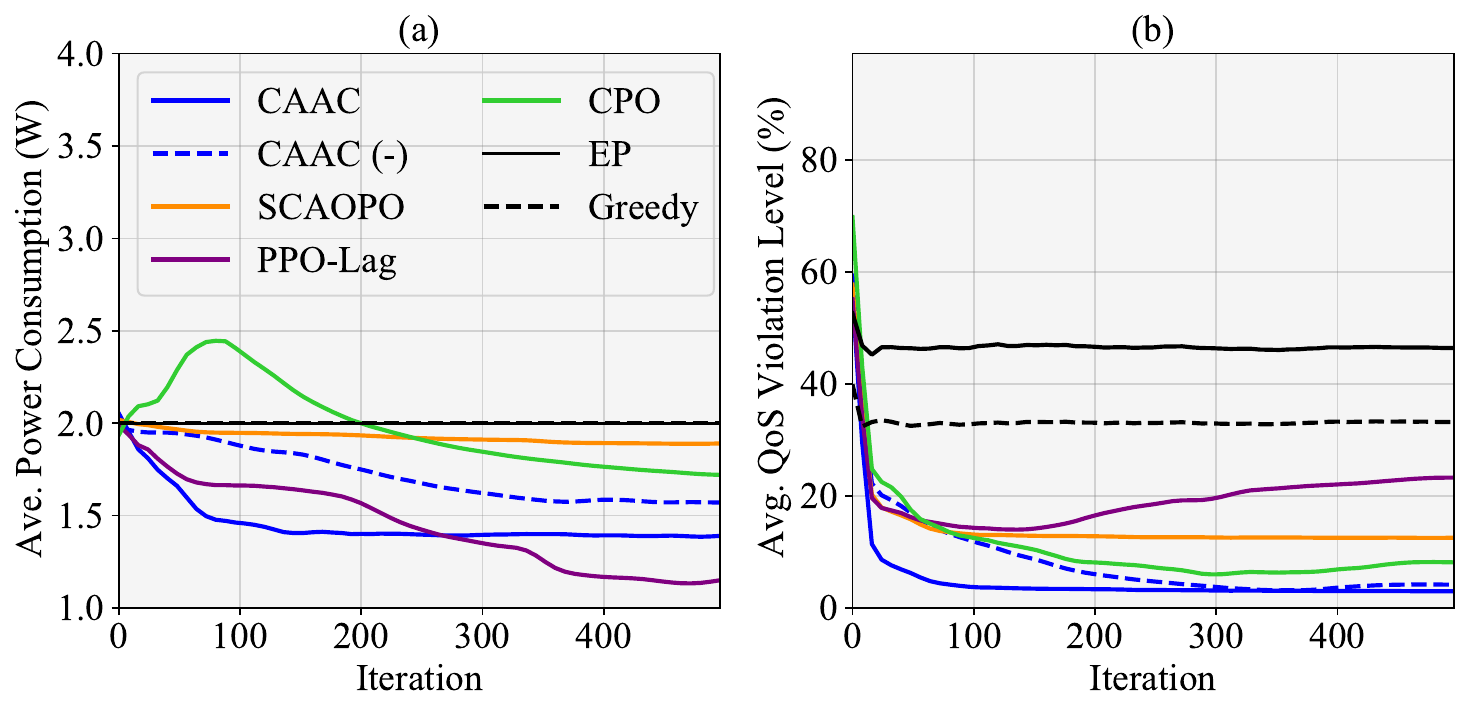}\vspace{-0.3cm}
\caption{\label{fig:convergence}Convergence curves of algorithms.}
\vspace{-0.5cm}
\end{figure}
\begin{figure}
\begin{centering}
\includegraphics[height=4.2cm]{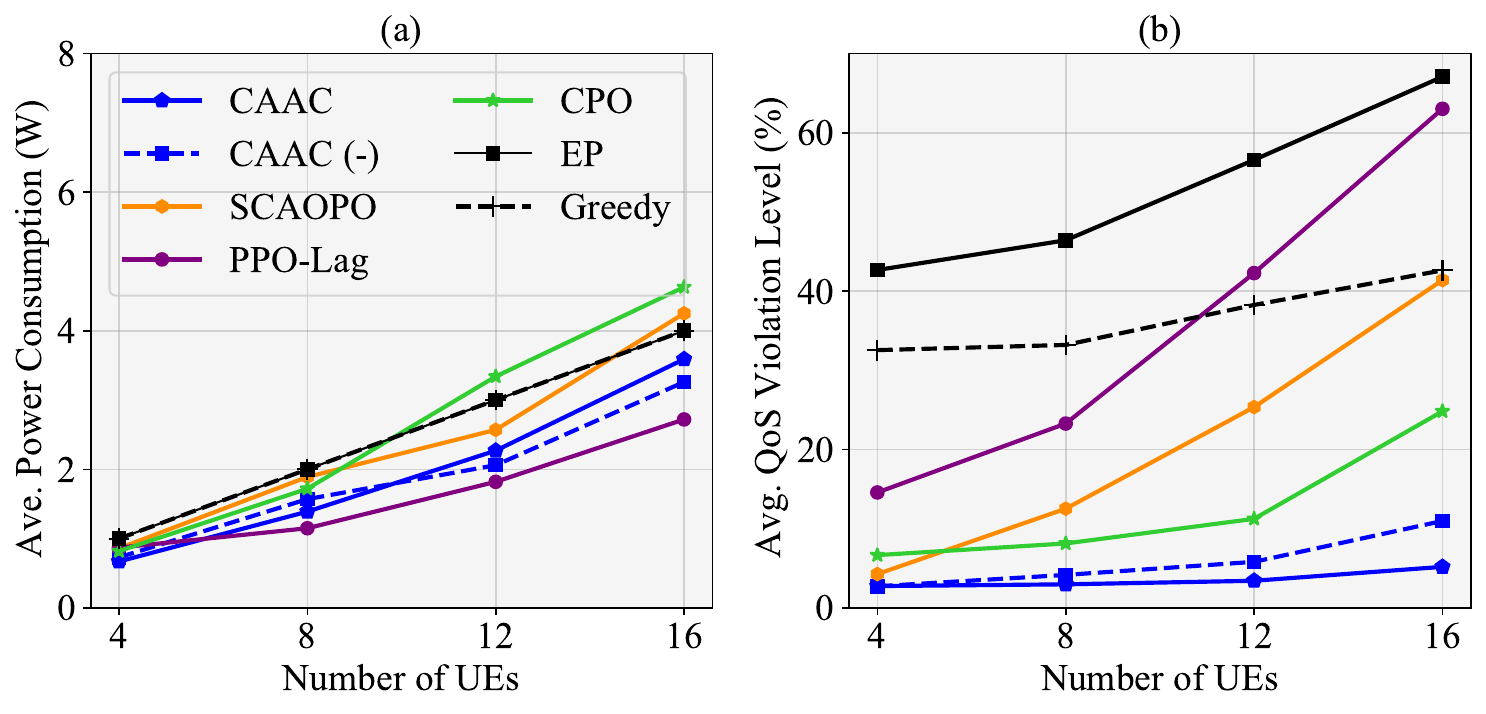}\vspace{-0.3cm}
\par\end{centering}
\caption{\label{fig:user}Performance versus the number of users.}
\vspace{-0.5cm}
\end{figure}
\begin{table*}
\caption{\label{tab:Comparison-of-Q-Network}Comparison of Q-Network Parameter
Sizes.}
\vspace{-0.3cm}

\centering{}{\small{}%
\begin{tabular}{|c|c|}
\hline 
{\small Algorithm} & {\small Parameterize Size}\tabularnewline
\hline 
\hline 
{\small Lightweight Attention-enhanced networks} & {\small$O\left(\left(2M+1\right)L_{1}K^{2}\left(K+1\right)+\left(L_{1}K+L_{3}\right)\left(L_{2}+K+1\right)\left(K+1\right)+L_{3}\left(K+1\right)\right)$}\tabularnewline
\hline 
{\small four-layer fully connected networks} & {\small$O\left(\left(2M+1\right)L_{1}K+\frac{3L_{1}L_{2}}{2}+\left(L_{2}+K+1\right)L_{3}\left(K+1\right)+L_{3}\left(K+1\right)\right)$}\tabularnewline
\hline 
\end{tabular}}\vspace{-0.5cm}
\end{table*}

Fig. \ref{fig:convergence} and Fig. \ref{fig:user} show algorithms'
convergence behavior under $K=8$ and their average performance over
500 iterations with varying user numbers, respectively. Specifically,
Fig. \ref{fig:convergence}-a and Fig. \ref{fig:user}-a depict the
average power consumption, while Fig. \ref{fig:convergence}-b and
Fig. \ref{fig:user}-b show the average QoS violation level across
users, which is measured by $\mathrm{QoS}{}_{\mathrm{gap}}\left(i\right)=\frac{1}{KB\left(i+1\right)}\stackrel[t=0]{B\left(i+1\right)-1}{\sum}\stackrel[k=1]{K}{\sum}\bigl[\stackrel[n=1]{N}{\sum}\delta\left(n-\varrho_{k}\right)\mathbb{U}\left(n,k,t\right)-c_{k}\bigr]^{+}$,
where $\left[\cdot\right]^{+}$ denotes the positive part operator.
Baselines include configured power control schemes with equal-priority
(EP) scheduling, a greedy priority scheduling that adjusts \cite{Qweighted}
using $\mathrm{QoS}{}_{\mathrm{gap}}\left(i\right)$ as priorities,
and three advanced Actor-Critic-based CRL algorithms (PPO-Lag, CPO,
and SCAOPO) introduced in Section \ref{sec:Introduction}. We also
design a CAAC variant, denoted as CAAC (-), by replacing the lightweight
attention-enhanced Q-networks with separate four-layer FCNs, where
the input and output dimensions of each layer in the latter match
those of the former. As shown in Fig. \ref{fig:convergence}, CAAC
exhibits fast and stable convergence, reducing the average power consumption
to 1.39 W and QoS violation level to 2.98\%. The greedy method slightly
outperforms the EP method, as it dynamically assigns higher priority
weights to users with consistently poor performance. However, both
methods are far from providing QoS guarantees to users, despite consuming
more power than CRL-based algorithms. CPO and PPO-Lag demonstrate
clear improvements over the greedy and EP methods. Nevertheless, as
discussed in Section \ref{sec:Introduction}, their policy optimization
methods are not well-suited to handling non-convex objectives and
constraints. As a result, CPO underperforms in power efficiency, while
PPO-Lag struggles to meet QoS requirements effectively. SCAOPO, though
effective in simpler scenarios in \cite{SCAOPO}, performs unsatisfactorily
in the more complex scenarios of this paper, as simple MC methods
struggle to achieve accurate policy evaluation. CAAC converges significantly
faster than CAAC (-), primarily because its parameter size is substantially
smaller than that of CAAC (-), as compared in Table. \ref{tab:Comparison-of-Q-Network}.
Moreover, CAAC achieves better final performance than CAAC (-), demonstrating
that the integration of the attention mechanism enhances the representational
capacity of the Q-networks. Fig. \ref{fig:user} shows CAAC consistently
outperforms EP and greedy methods across all user scenarios. Although
CRL-based algorithms perform similarly when there are fewer users,
their performance gap with CAAC grows as the number of users increases.
This is because large-scale user scenarios demand higher learning
efficiency of Q-networks and make the problem (\ref{eq:original-problem})
more non-convex, further amplifying the challenge for policy optimization.
The advanced Q-network architecture and CSSCA-based policy optimization
method are well-suited to address these requirements and challenges.

\section{Conclusion}

This paper proposes the CAAC algorithm for adaptive UPTPS in WMMSE
precoding, aiming to minimize energy consumption while meeting heterogeneous
QoS constraints. CAAC consists of an Actor module and a Critic module.
The Actor executes the scheduling policy and optimizes it using a
novel CSSCA method, which effectively handles non-convex stochastic
objectives and QoS constraints. Meanwhile, the Critic evaluates the
policy using lightweight attention-enhanced Q-networks, which are
trained in a model-free manner and exhibit high learning efficiency.
Simulation results demonstrate that CAAC offers superior energy efficiency
and QoS satisfaction, making it a promising solution for green and
intelligent resource allocation in future wireless systems.

\vspace{-0.3cm}
\bibliographystyle{IEEEtran}
\addcontentsline{toc}{section}{\refname}\bibliography{RLreferences}

\end{document}